\documentclass[letterpaper]{article} 
\usepackage{aaai2026}  
\usepackage{times}  
\usepackage{helvet}  
\usepackage{courier}  
\usepackage[hyphens]{url}  
\usepackage{graphicx} 
\usepackage{natbib}  
\usepackage{caption} 
\usepackage{algorithm}
\usepackage{algorithmic}
\usepackage{amsmath, amsfonts}
\usepackage{multirow}
\usepackage{booktabs} 
\usepackage{subcaption}
\usepackage{xspace}

\usepackage{newfloat}
\usepackage{listings}
\DeclareCaptionStyle{ruled}{labelfont=normalfont,labelsep=colon,strut=off} 
\floatstyle{ruled}
\newfloat{listing}{tb}{lst}{}
\floatname{listing}{Listing}
\title{Self-Supervised Learning Based on Transformed Image Reconstruction for Equivariance-Coherent Feature Representation}
\author{
    Qin Wang\textsuperscript{\rm 1,\rm 3},
    Alessio Quercia\textsuperscript{\rm 1,\rm 3},
    Benjamin Bruns\textsuperscript{\rm 1}, \\
    Abigail Morrison\textsuperscript{\rm 2,\rm 3}, 
    Hanno Scharr\textsuperscript{\rm 1},
    Kai Krajsek\textsuperscript{\rm 4}
}
\affiliations{
    \textsuperscript{\rm 1}Institute for Advanced Simulation (IAS-8): Data Analytics and Machine Learning\\
    Forschungszentrum Jülich GmbH, Jülich, Germany\\
    \textsuperscript{\rm 2}Institute for Advanced Simulation (IAS-6): Computational and Systems Neuroscience\\ Forschungszentrum Jülich GmbH, Jülich, Germany \\
    \textsuperscript{\rm 3}Software Engineering, Faculty of Computer Science, RWTH Aachen University, Aachen, Germany\\
    \textsuperscript{\rm 4}Jülich Supercomputing Centre (JSC), Forschungszentrum Jülich GmbH, Jülich, Germany \\
    \{qi.wang, a.quercia, b.bruns, a.morrison, h.scharr, k.krajsek\}@fz-juelich.de
}

\begin{document}

\maketitle

\begin{abstract}
Self-supervised learning (SSL) methods have achieved remarkable success in learning image representations allowing invariances in them — but therefore discarding transformation information that some computer vision tasks actually require. While recent approaches attempt to address this limitation by learning equivariant features using linear operators in feature space, they impose restrictive assumptions that constrain flexibility and generalization.
We introduce a weaker definition for the transformation relation between image and feature space denoted as equivariance-coherence. We propose a novel SSL auxillary task that learns equivariance-coherent representations through intermediate transformation reconstruction, which can be integrated with existing joint embedding SSL methods. Our key idea is to reconstruct images at intermediate points along transformation paths, e.g.\ when training on 30° rotations, we reconstruct the 10° and 20° rotation states. 
Reconstructing intermediate states requires the transformation information used in augmentations, rather than suppressing it, and therefore fosters features containing  the augmented transformation information.
Our method decomposes feature vectors into invariant and equivariant parts, training them with standard SSL losses and reconstruction losses, respectively. We demonstrate substantial improvements on synthetic equivariance benchmarks while maintaining competitive performance on downstream tasks requiring invariant representations. The approach seamlessly integrates with existing SSL methods (iBOT, DINOv2) and consistently enhances performance across diverse tasks, including segmentation, detection, depth estimation, and video dense prediction. Our framework provides a practical way for augmenting SSL methods with equivariant capabilities while preserving invariant performance.
\end{abstract}

\section{Introduction}
\label{sec:intro}

\begin{figure*}[t]
    \centering
    \includegraphics[width=0.95\linewidth]{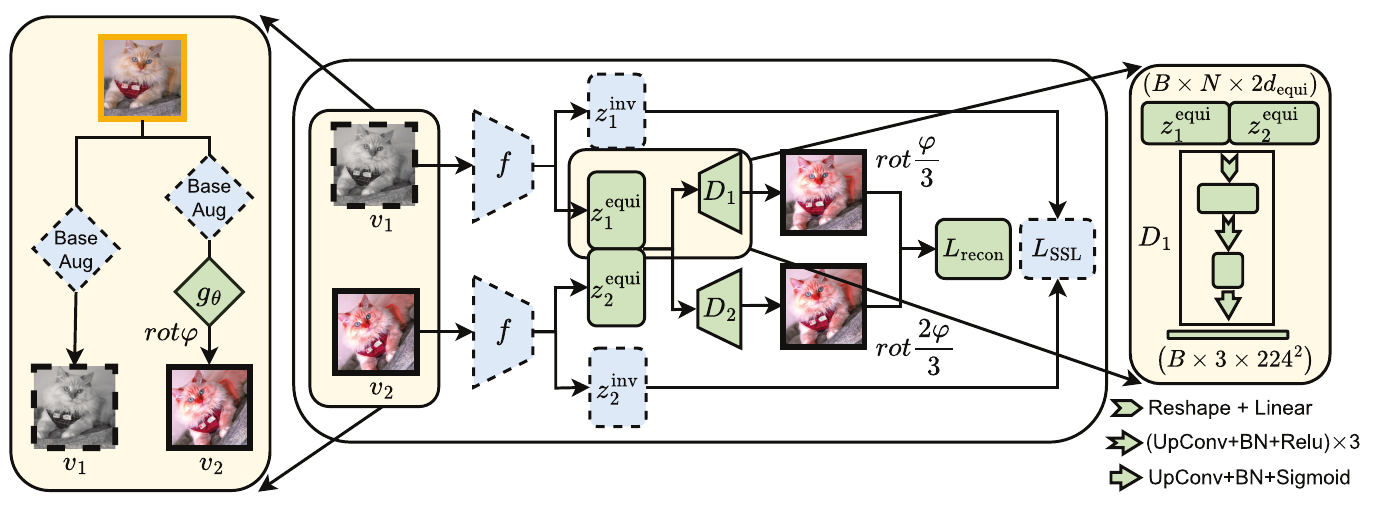}
\caption{Overview of our framework for equivariance-coherent feature representation. As done in many joint embedding SSL methods like DINOv2, we apply two sets of image augmentations generating two different views of the original image. On the secondary view, we additionally apply a sequence of transformations $g_i$ (e.g., three rotations by the angles $\frac{\varphi}{3}$,$\frac{2\varphi}{3}$ and $\varphi$ as shown in the figure) to enforce equivariant structure during transformed image reconstruction.The last rotation serves as the second view for the joint embedding whereas the in-between transformed images are the targets of the reconstruction task.}
\label{fig:method_overview}
\end{figure*}

Self-supervised learning (SSL) methods \cite{chen2020simple,he2020momentum, chen2020mocov2, mae, i-jepa, Oquab2023DINOv2LR, ibot} have become crucial for pretraining foundation models by leveraging unannotated images for representation learning. However, joint embedding SSL methods, currently one of the best performing SSL methods,  face a fundamental trade-off: they are designed around surrogate tasks that promote invariance, i.e.\ learning features that remain unchanged under input transformations. 
This invariance bias is reinforced due to the most popular evaluation being linear probing on ImageNet-1K \cite{imagenet}, where classification labels are inherently invariant to transformations used in data augmentation.

While invariance is valuable for classification tasks, many computer vision applications require equivariant features that preserve transformation information rather than discarding it. 
Equivariance means that when an input undergoes a transformation (e.g.\ rotation, translation, scaling), the learned representation changes in a predictable, recoverable way. 
This property is essential for dense prediction tasks like object detection, segmentation, or pose estimation, where knowing precise object positions and orientations is critical for understanding where objects are in a scene. Recent methods like the Split Invariant–Equivariant (SIE) framework \cite{garrido2023sie} attempt to address this limitation by learning both invariant and equivariant features through transformation-conditioned linear operators in latent space. While effective, SIE imposes restrictive architectural constraints and assumes equivariant mappings to be linear. Such assumptions may be too strict for complex transformations or non-rigid deformations.

Our key idea is to use augmentation information in an auxiliary task, rather than modelling equivariance in latent space. 

Our method reconstructs images at intermediate points along transformation paths. For example, when training on a 30° rotation, we reconstruct images at 10° and 20° rotation angles. 
This intermediate reconstruction pushes the model to provide features containing information on the transformation and thus encourages equivariant learning. 
Compared to SIE, our approach offers two key advantages: (1) it removes linearity constraints on equivariant mappings, broadening the function space for learning unseen or complex transformations, (2) it simplifies architecture by avoiding hyper-networks and operator prediction modules.

Empirical results in Table \ref{tab:r2_comparison} demonstrate superior performance of our reconstruction-based approach across multiple benchmarks, with strong generalization capabilities that transfer effectively to unseen transformations. Our contributions are:

\begin{table*}[tb]
    \centering
    \small
    \begin{tabular}{ l c c c c c}
        \hline
        \textbf{Configurations} & $R^2(\text{rot})$ & $R^2(\text{color})$ & $R^2(\text{blur})$ & $R^2(\text{trans})$ & Mean($R^2$)\\
        \hline
        SIE(rot)  & 0.990 & 0.867 & 0.042 & 0.540  & 0.610 \\ 
        SIE(color) & 0.078 & 0.890 & 0.097 & 0.355 & 0.355  \\ 
        SIE(blur)  & 0.153 & 0.883 & 0.941 & 0.189 & 0.542  \\ 
        SIE(trans) & 0.213 & 0.885 & 0.023 & 0.978 & 0.525 \\ 
        SIE(all) & 0.331 $\pm$ 0.007 & 0.899 $\pm$ 0.003 & 0.211 $\pm$ 0.005 & 0.925 $\pm$ 0.002 &0.592\\
        Cross Atten Recon \shortcite{wang2024equivariant} & 0.893 $\pm$ 0.004 & 0.921 $\pm$ 0.006 & 0.823 $\pm$ 0.030 & 0.875 $\pm$ 0.005& 0.878\\
        \hline
        Ours(VICReg, rot, rot) & 0.9975 $\pm$ 0.0005 & 0.9073 $\pm$ 0.0021 & 0.9310 $\pm$ 0.0020 & 0.9810 $\pm$ 0.0010 &0.954\\
        Ours(VICReg, all, rot) & \textbf{0.9983 $\pm$ 0.0005} & 0.9231 $\pm$ 0.0005 & 0.9689 $\pm$ 0.0099 & 0.9801 $\pm$ 0.0004 &0.968 \\
        Ours(VICReg, all, color) & 0.9891 $\pm$ 0.0019 & \textbf{0.9373 $\pm$ 0.0013} & 0.9700 $\pm$ 0.0067 & 0.9699 $\pm$ 0.0022 &0.967 \\
        Ours(VICReg, all, blur) & 0.9981 $\pm$ 0.0001 & 0.9154 $\pm$ 0.0006 & 0.9392  $\pm$ 0.0106 & 0.9774 $\pm$ 0.0007 &0.958\\
        Ours(VICReg, all, trans) & 0.9975 $\pm$ 0.0005 & 0.9288 $\pm$ 0.0012 & \textbf{0.9747 $\pm$ 0.0017} & \textbf{0.9830 $\pm$ 0.0004} &\textbf{0.971}\\
        Ours(VICReg, all, SE(2)) &0.9980 $\pm$ 0.0001 & 0.9158 $\pm$ 0.0005 & 0.9520 $\pm$ 0.0050 & 0.9740 $\pm$ 0.0001 &0.960\\
        \hline
    \end{tabular}
    \caption{Comparison of $R^2$ values across different configurations for synthetic tasks. Configuration naming: SIE methods show training transformation in brackets. 
    Our methods use format: Ours(SSL loss, augmentation, reconstruction target) where 
    `SSL loss' is the $\mathcal{L}_{SSL}$ function used, `augmentation' is the DINOv2 common augmentation pipeline 
    (rot = rotation only, all = all common augmentations), and `reconstruction target' is the 
    target transformation for $\mathcal{L}_\text{recon}$. \textbf{Bold} values indicate overall best. We use VICReg \cite{bardes2022vicreg} as the invariance loss, consistent with SIE's approach. }
    \label{tab:r2_comparison}
\end{table*}

\begin{itemize}
    \item \textbf{New task for joint invariant-equivariant learning}: We introduce a reconstruction-based task that learns generalised equivariant features, relaxing restrictive assumptions about linear equivariant mappings. 
    \item \textbf{Strong empirical validation on synthetic benchmarks}: Our method achieves superior performance on all synthetic equivariance tasks from \cite{wang2024equivariant}, demonstrating clear advantages over existing approaches including SIE.

    \item \textbf{Consistent improvements across diverse real-world tasks}: Compared to strong baselines (iBOT \cite{ibot}, DINOv2 \cite{Oquab2023DINOv2LR}), our approach improves performance on most evaluation tasks while maintaining competitive results on  invariance benchmarks.

    \item \textbf{General framework compatible with existing SSL methods}: Our intermediate reconstruction approach can be integrated with various SSL frameworks, allowing to enhance existing methods with equivariant capabilities.
\end{itemize}

\section{Related Work}
\label{sec:related_work}

\subsection{Equivariant Neural Networks}

Since the early days of neural networks research, the exploration of symmetries in the data has played a significant role, reduced model complexity, and improved inference quality of models \cite{fukushima:neocognitronbc}. One might argue that without convolutional neural networks, inherently implementing approximately translational equivariance, computer vision models could not have made this progress in the field. However, built-in permutation equivariance in transformer architectures has also been the object of study \cite{Xu2023PermutationEO}.  
%
Equivariance in deep learning can be split in two
sub-categories: studies on models that inherit built-in equivariance \cite{cohen2016steerablecnns,jenner2022steerable} or models that gain this property by experience \cite{Xiao2020WhatSN,Dangovski2021EquivariantCL,Wang2024UnderstandingTR}.

\subsection{Self-Supervised Learning}

State-of-the-art SSL methods learn feature representations by automatically labelling non-labeled data and applying supervised learning techniques. The assumption is that the learned feature representation is comprehensive enough to be used later in other tasks, denoted as downstream tasks. A large variety of different SSL methods have been proposed \cite{he2020momentum,chen2020simple,He22,chen2020mocov2,zbontar2021barlow,caron2021emerging,bardes2022vicreg,lehner2023contrastive,Xie2024SelfGuidedMA,Oquab2023DINOv2LR}.
For an overview, we refer to \cite{balestriero2023cookbook}. 

For our purpose it is relevant, how models react to different transformations in the input space, i.e.\ if they maintain the information in the feature representation or if this information is suppressed. 
Older SSL methods proposed auxiliary tasks such as Jigsaw puzzle \cite{Noroozi2016UnsupervisedLO} or rotation estimation \cite{Gidaris2018UnsupervisedRL} that foster equivariance properties as the transformation properties need to be maintained in feature space. These methods have been overtaken by matching-type methods. They present different versions of the same semantic content to the model and motivate it to map them to nearby points in feature space. Semantically different images are mapped to points far apart. This is achieved either by contrastive learning approaches or by regularisation techniques like de-correlation \cite{zbontar2021barlow,bardes2022vicreg} or teacher-student architectures \cite{ibot,Oquab2023DINOv2LR}. Consequently, these matching-type methods learn feature representations that suppress the differences between the versions.

Another state-of-the-art SSL branche includes mask-based approaches that remove parts of the input image and reconstruct them or a transformed version of them, either in the original image space \cite{He22,Bandara2022AdaMAEAM}, or in feature space \cite{i-jepa}. Contrastive learning has been combined with masked approaches, but only pixel-accurate translation has been applied as augmentation \cite{Huang2022ContrastiveMA}.  As these methods do not, apart from masking or cropping, rely on other transformations, they are by construction more open for equivariance.
Our approach is closely related to SIE \cite{garrido2023sie}, combining the matching approach with an explicit model of transformations applied in the input space. Our approach can be seen as an extension not requiring knowledge about the transformation parameters.  

In contrast to SIE, our approach reaches state-of-the-art results.    

\subsection{Equivariance vs.\ Invariance}

A function $f$ is denoted as equivariant with respect to a transformation $t$ in the input space and a corresponding transformation $\hat{t}$ in the output space, if the function commutes with the transformations, i.e.\ $\hat{t}(f(x)) = f(t(x))$. Here, $f$ is a deep learning model, input space is the space of images or videos, and output space is the feature space.

The definition includes the identity transformation in the output space such that invariance is always also equivariance. However, in the computer vision literature e.g.\  \cite{Xiao2020WhatSN,Dangovski2021EquivariantCL,Devillers2022EquiModAE,Garrido2023SelfsupervisedLO,Park2022LearningSE,Gupta2023StructuringRG,Wang2024UnderstandingTR}, as we do in this paper,  invariance is often opposed to equivariance to stress that all information about the transformation in the input space is still retrievable from the output space. The term equivariance is usually considered for a set of transformations, i.e.\  the function is said to be equivariant with respect to this set of transformations. Moreover, the set of transformations is considered a group transformation or, even stronger, a group representation of the transformation, i.e.\ a linear map,  in the input and output space. However, not all transformations applied in computer vision can be modelled by group transformations like elastic distortions, crop-resize operations, or non-affine perspective transformations. In addition, transformations that can theoretically be formulated as group transformations might lose the corresponding group properties in practice. E.g.\ an image rotated around a general angle cannot be rotated back as parts of the image get lost during the forward transformation. In this paper, we do not restrict the model to be equivariant with transformations that belong to a certain further structure. We argue that equivariance is no value in itself but should serve as a property to help to learn a feature representation that contains all the information necessary for all kinds of downstream tasks. We do not restrict the transformation to form a group or require them to be linear in the feature space. Instead we motivate the model to maintain the information that is necessary to maintain input output relation such that transformed images in the input space can be retrieved by the representation. It shall be irrelevant if the transformation forms a linear transformation,  a general group transformation, or even if the definition of equivariance is only approximately fulfilled in the feature representation. We denote this as equivariant-coherence.

\section{Method}

Our approach extends SSL frameworks with an auxiliary reconstruction task that learns equivariance-coherent features. The key innovation is intermediate reconstruction: rather than learning from just the original and final transformed images, we supervise the model to reconstruct images at multiple points along transformation trajectories. This design naturally encourages the model to retain information necessary to perform the considered transformations. Given transformation $g_\theta$ defined by a parameter vector $\theta=[\varphi; t_x; t_y; ...]$ of continuous parameters,   we define $K$ intermediate transformations $g_{\theta_1}, g_{\theta_2}, \ldots, g_{\theta_K}$ where $\theta_k:=\frac{k\theta}{K+1}$ for $k \in \{1,2,\ldots,K\}$ are parameter vectors consisting of equidistant parameters values between no transformation besides the base transformation  and the additional transformation defined by $\theta$. In a first step to generate views as input for our joint embedding SSL approach a first view $v_1=\mathcal{A}_1(I)$ is generated by means of a set $\mathcal{A}_1$ of augmentation transformations. The second view $v_2$ is generated by a second set $\mathcal{A}_2$ of augmentation  transformations $u:=\mathcal{A}_2(I)$ and, in contrast  to the first view, it undergoes a sequence of additional  transformations (the transformations are listed in Table \ref{tab:group_transformations}) 
\begin{equation}
  u \to g_{\theta_1}(u) \to g_{\theta_2}(u) \to \cdots \to g_{\theta_K}(u) \to g_\theta(u)  
  \label{eq:g_theta}
\end{equation}
where the end of the sequence constitutes the second view $v_2:=g_\theta(\mathcal{A}_2(I))$ in our joint embedding SSL method.  

Intermediate images $u_k := g_{\theta_k}(u)$ are to be reconstructed by the SSL method acting  as anchor points shaping the geometry of the feature space. 
We empirically determine that $K = 2$ 
yields optimal performance (Table \ref{tab:num_images}).

\subsection{Feature Splitting}

In our proposed framework, depicted in Figure~\ref{fig:method_overview}, we introduce transformation $g$ as an additional augmentation applied exclusively to $u\in \mathbb{R}^{B\times 3 \times 224 \times 224}$ where $B$ denotes the batch size. Both views, $v_1$ and $v_2$,  are processed through encoder $f$, which operates either with shared weights or within a student-teacher framework depending on the SSL method used. In the student-teacher setup, the teacher network is updated using an exponential moving average (EMA) of the student's parameters. 

The resulting feature representations $z_i \in \mathbb{R}^{B \times N \times d_{\text{patch}}}$, $i \in \{1,2\}$, are split into two complementary components along the feature dimension, where $d_{\text{patch}} = d_{\text{inv}} + d_{\text{equi}}$, $N$ is the number of patches, and $d_{\text{patch}}$ is the patch embedding dimension:

\begin{itemize}
    \item \textbf{Invariant features} $z_i^{\text{inv}} \in \mathbb{R}^{B\times N\times d_{\text{inv}}}$  are supervised using standard SSL losses, 
    e.g., the iBOT loss.
    \item \textbf{Equivariant features} $z^{\text{equi}}_i \in \mathbb{R}^{B\times N\times d_{\text{equi}}}$  are used to reconstruct intermediate transformed images.
\end{itemize}
The dimension of the equivariant feature component, denoted as $d_{\text{equi}}$, can be adjusted as needed. We perform sensitivity analyses on $d_{\text{equi}}$ to examine its influence.

\subsection{Intermediate Transformation Reconstruction}
We employ $K$ independent decoders, $D_1, D_2, \ldots, D_K$, that operate on the concatenated equivariant features $[z^{\text{equi}}_1; z^{\text{equi}}_2]$ to reconstruct intermediate transformed versions of the input image.
To reduce computational cost and emphasize the encoder's role during pretraining, we deliberately design simple decoders consisting of a single linear layer followed by four convolutional layers (see Figure~\ref{fig:method_overview}). Our objective is not to achieve high-fidelity reconstruction, but rather to provide sufficient supervisory signal for the encoder $f$ to learn effective equivariant representations. 
The lightweight decoder design ensures that the primary learning occurs in the encoder without interference from complex reconstruction architectures.
Each decoder produces reconstructions:
\begin{equation}
\hat{u}_k = D_k([z^{\text{equi}}_1; z^{\text{equi}}_2]) \quad \text{for } k = 1, 2, \ldots, K
\end{equation}
The reconstruction loss $\mathcal{L}_{\text{recon}}$ is computed as the mean $L_2$ loss between each decoder's prediction $\hat{u}_k$ and the corresponding ground truth intermediate images $u_k$
\begin{equation}
\mathcal{L}_{\text{recon}} = \frac{1}{K} \sum_{k=1}^{K} \|\hat{u}_k - u_k\|_2^2
\end{equation}
This reconstruction objective is combined with the standard SSL loss $\mathcal{L}_{\text{SSL}}$ using a weighting hyperparameter $\lambda$:
\begin{equation}
\mathcal{L}_{\text{total}} = \mathcal{L}_{\text{SSL}}(z_1^{\text{inv}}, z_2^{\text{inv}}) + \lambda  \mathcal{L}_{\text{recon}}
\label{eq:total_loss}
\end{equation}
%
A sensitivity analysis for 
$\lambda$ and $d_{\text{equi}}$ is given in Section~\ref{subsec:sensitivity-analysis}.

\subsection{Transformation Types and Parameters} 

We introduce group transformations, approximate group transformation and non-group transformations in the reconstruction process to learn equivariant features. These group transformations include geometric transformations such as rotation, translation, and special Euclidean group transformations SE(2). 
\begin{table}[t]
\small
    \centering
        \begin{tabular}{c c c }
        \hline
        \textbf{Transformations $g$} & \textbf{Parameters} &\textbf{Mag. range} \\
        \hline
        Rotation & angle $\varphi$ & $[-45, 45]$\\
        \hline
       Color jittering & \multicolumn{1}{c}{brightness, contrast,} & $[-0.4, 0.4]$ \\
         & \multicolumn{1}{c}{saturation $S$, hue $H$} &$[-0.1, 0.1]$ \\
        \hline
        Gaussian blur & radius $\sigma$ & $[0.1, 2]$ \\
        \hline
        Translation & displacement $t_x, t_y$& $[-10, 10]$ \\
        \hline
        SE(2) &\multicolumn{1}{c}{angle $\varphi$} & $[-45, 45]$ \\
         & \multicolumn{1}{c}{displacement $t_x, t_y$} &$[-10, 10]$ \\
        \hline
        \end{tabular}
    \caption{Considered transformations to generate the intermediate transformed images used as targets of the auxiliary  reconstruction task, as well as the second view for the joint  embedding.} 
\label{tab:group_transformations}
\end{table}

SE(2) defines the isometries in $\mathbb{R}^2$ that preserve the orientation, i.e.\ it combines 2d rotation and translation. 

Additionally, we incorporate non-geometric transformations, including color jittering and Gaussian blur. Transformation parameter ranges are shown in Table \ref{tab:group_transformations}.

\section{Experimental Results}

As our method builds on SIE \cite{garrido2023sie}, SIE is our first natural baseline with synthetic evaluation tasks designed to benefit from equivariant features. 
To evaluate the generalization of our method against state-of-the-art approaches, we go beyond using the invariant loss function $L_\text{SSL}$ as in SIE and explore integrating other methods. Specifically, we incorporate co-learning with iBOT \cite{ibot} and DINOv2 \cite{Oquab2023DINOv2LR}, both 
augmentation-based techniques. They are also tested on the synthetic benchmark.
Finally, we evaluate on a rich set of more realistic downstream tasks. We aim to enhance the baselines' performance on equivariance-related tasks while preserving strong results on invariance-related benchmarks, e.g.\ ImageNet linear probing.

\subsection{Implementation Details}

\subsubsection{Architecture}
We use Vision Transformers (ViTs) \cite{vit} with different configurations, specifically ViT-S/16 and ViT-L/16, as backbones for experiments. We incorporate a linear head on top of the backbone as originally done by the baseline methods to accommodate different representation dimensions $d_{\text{patch}}$, i.e.\ 8192 for iBOT, 512 for SIE, and 2048 for DINOv2.  
 
Afterwards, a portion $z^\text{equi}$ of these features $z$ is allocated for reconstruction. We do not introduce more features than the baseline methods.


\subsubsection{Pretraining Setup}
Our approach uses a baseline SSL loss $L_\text{SSL}$ in addition to our new component $L_\text{recon}$. Each of the three baseline methods come with distinct training setups. The common training configuration includes ImageNet-1K as the dataset, optimizer AdamW \cite{adamw}, and a cosine-scheduled learning rate.  For the SIE-based method, we apply their invariant loss as $L_{SSL}$, and pretrain ViT-S/16 for 800 epochs with a batch size of 2048. The base learning rate is set to $ 10^{-4} $ and is linearly scaled with the batch size $B$: $lr = 10^{-4} \cdot B/256$. For the iBOT-based approach, we pretrain ViT-S/16 for 800 epochs and ViT-L/16 for 250 epochs, both with a batch size of 1024. The learning rate follows the same linear scaling strategy, with a base learning rate of $ 5e{-4} $. For the DINOv2-based training, we train ViT-L/16 with 100 epochs with batch size of 2048. The base learning rate is $4e{-3}$ with warmup 10 epochs. In the pretraining stage, we weight the reconstruction loss introduced by the auxiliary task using the optimal coefficient $\lambda = 1$, as determined by the sensitivity analysis in Section~\ref{subsec:sensitivity-analysis}. For the equivariant feature dimension $d_{\textit{equiv}}$, we also select the hyperparameter based on this sensitivity analysis, adopting a default value of $2048$. For DINOv2, we retain the same proportional relationship between feature dimensions and therefore set $d_{\textit{equiv}} = 512$. For VICReg, we follow SIE and evenly split the embedding dimension to determine $d_{\textit{equiv}}$.

\subsubsection{Computation Cost}
We conduct all pretraining experiments on the JUWELS Booster \cite{JUWELS}. For VICReg, we adapt the method to a ViT-S/16 backbone and train on 4 nodes with a total of 16 A100 GPUs. Both iBOT and DINOv2 are pretrained on 16 nodes with 64 A100 GPUs. The computational cost of the different SSL methods, along with the additional cost introduced by our auxiliary tasks, is summarized in Table \ref{tab:overhead_comparison}. Overall, our lightweight intermediate-transformation reconstruction task adds only a modest overhead to the baseline SSL methods.
\begin{table}[hb]
\small
    \centering
        \begin{tabular}{l c c c}
        \hline
        \textbf{Config.} & Backbone & Runtime s/epoch & Overhead  \\
        \hline
        VICReg & ViT-S/16 & 156 & -\\
        + Ours & ViT-S/16 & 179& 14.7\%\\
        \hline
        iBOT & ViT-L/16 & 230 & - \\
        + Ours & ViT-L/16& 249 & 8.3\% \\
        \hline
        DINOv2 & ViT-L/16 & 750 & - \\
        + Ours & ViT-L/16 & 804 & 7.2\% \\
        \hline
        \end{tabular}
    \caption{Computational overhead, where Ours refer to Ours(all, SE(2)), with all base augmentations and SE(2) transformation.}
\label{tab:overhead_comparison}
\end{table}
\begin{figure}[tb]
    \centering
    \includegraphics[width=0.9\linewidth]{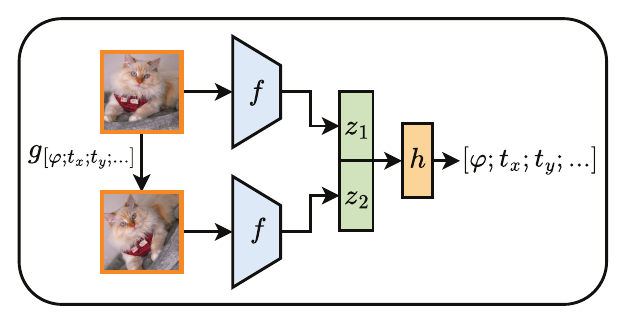}
    \caption{Synthetic tasks for evaluating equivariant representations. Transformation ($g$) is applied to the original image $I$. Both the transformed and original images are processed through a pretrained encoder $f$. A lightweight MLP $h$ then predicts the parameters of the applied transformation.}
    \label{fig:sythetic_tasks}
\end{figure}

\subsection{Performance on Synthetic Tasks}
\label{subsec:comparison_synthetic_tasks}

Following \cite{wang2024equivariant}, we design synthetic tasks to evaluate the equivariant representations learned during pretraining, see Figure~\ref{fig:sythetic_tasks}. These regression-based tasks assess the transformation parameters between original and transformed views. Following the evaluation metrics used in SIE \cite{garrido2023sie}, we formulate transformation parameters prediction as a regression problem. To quantify alignment between predicted and true values, we use the coefficient of determination ($R^2 = 1 - \frac{RSS}{TSS}$), where $RSS$ is the sum of squared residuals, and $TSS$ is the total sum of squares. A higher $R^2$ value indicates more accurate transformation predictions.

\subsubsection{Comparison with Other SIE-Based Methods}
Here, we use ViT-S/16 for all tests. Our approach adopts the SIE configuration and leverages VICReg~\cite{bardes2022vicreg} as $L_\text{SSL}$, consistent with SIE~\cite{garrido2023sie}. We compare with SIE  and a closely related cross-attention-based reconstruction method \cite{wang2024equivariant}, see Table \ref{tab:r2_comparison}.
For SIE, the performance with single augmentations and prior knowledge is the best among the baseline methods, but for color jitter estimation, where all variants perform well. However, it generalizes poorly to other transformation evaluations (best mean $R^2 = 0.610$). 
Cross-attention reconstruction \cite{wang2024equivariant} leads to much more balanced results (mean $R^2 = 0.878$).
For our method, we tested different combinations of augmentation and intermediate transformation. All of them show a strong average performance increase (mean $R^2$ from 0.954 to 0.971).

The best ones, i.e.\ Ours(VICReg, all, rot) and Ours(VICReg, all, trans), outperform all competitors on all tasks individually, demonstrating the most versatile equivariant representations.

\subsubsection{Enhancing Transformation Prediction of Augmentation-Based SSL Methods}
We test pretrained state-of-the-art ViT-L/16 models on the same rotation and color jitter prediction task (as outlined in \ref{subsec:comparison_synthetic_tasks}), see Table~\ref{tab:performance_comparison_ROT_COLOR}. The baseline iBOT \cite{ibot} pretrained model is taken from the official repository. For fair comparison, numbers shown for the DINOv2 baseline model are for a model we pretrained from scratch, as it performed better than with the weights from the official repository. For color prediction, iBOT and DINOv2 perform on par with the smaller ViT-S/16 models from Table~\ref{tab:r2_comparison}. However, DINOv2 and iBOT perform significantly worse on the rotation estimation task. Training them from scratch using our approach improves their performance to best-in-class for color prediction (Ours(iBOT, all, SE(2)). For rotation prediction, improvements for iBOT are remarkably high from 0.238 to 0.937. DINOv2 performance is improved less and to a lower performance (0.84). Notably, using SE(2) as intermediate transform for iBOT and DINOv2 works slightly better than rotation, w.r.t. SIE, see Table~\ref{tab:r2_comparison}.

\begin{table}[tb]
\small
    \centering
        \begin{tabular}{l c c c }
        \hline
        \textbf{Configuration} &  $R^2(\text{rot})$ &  $R^2(\text{color})$ &  Mean($R^2$)\\
        \hline
        SimCLR & 0.288 & 0.575 & 0.432 \\
        + Ours(all, rot) & 0.637&0.712 & 0.675\\
        + Ours(all, SE(2)) & 0.705 & 0.723 & 0.714 \\
        \hline
        IBOT & 0.238 & 0.913 & 0.576 \\
        + Ours(all, rot) & 0.925 & 0.934 & 0.930 \\
        + Ours(all, SE(2)) &\textbf{ 0.937} & \textbf{0.943} & \textbf{0.940 }\\
        \hline
        DINOv2 & 0.774 & 0.910 & 0.842 \\
        + Ours(all, rot) & 0.812 & 0.920 & 0.866 \\
        + Ours(all, SE(2)) & 0.840 & 0.933 & 0.887 \\
        \hline
        \end{tabular}
    \caption{Performance comparison on rotation and color prediction tasks with improvements of our methods (absolute values). See Table~\ref{tab:r2_comparison} for description of configuration names.}
\label{tab:performance_comparison_ROT_COLOR}
\end{table}

\subsection{Performance on Natural Images Tasks}

We explore out method's impact on real-world imaging tasks commonly studied in self-supervised learning (SSL). We aim to 
be on par with state-of-the-art approaches, when downstream tasks are not to be expected to benefit from equivariance, like classification tasks. We strive to identify tasks where equivariant features are particularly beneficial.

Unless said differently, we use our method with iBOT and DINOv2 configurations performing best on the synthetic tasks from Table~\ref{tab:performance_comparison_ROT_COLOR}, i.e.\ Ours(iBOT, all, SE(2)) and Ours(DINOv2, all, SE(2)). Below, we call them Ours(iBOT) and Ours(DINOv2), respectively.

\begin{table*}[th]
    \centering
    \small
    \begin{tabular}{l c c c c c c c c}
        \hline
        \textbf{Configuration or Method} & \textbf{CIFAR10} & \textbf{CIFAR100} & \textbf{Aircraft} & \textbf{Pet} & \textbf{Food} & \textbf{Flowers}& \textbf{INat18} &\textbf{ImageNet} \\
        \hline
        iBOT & 97.60 & 86.96 & 55.43 & 92.30 & 88.39 &90.64 &57.30&81.00 \\
        +Ours(all, SE(2)) & 98.08 & 87.36 & 57.55 & 94.34 &\textbf{88.66} &96.03&57.99& 81.44 \\
        DINOv2&98.47&89.28&70.89&94.82&87.92&96.39&69.42&82.60 \\
        +Ours(all, SE(2))&\textbf{98.91}&\textbf{90.37}&\textbf{71.64}&\textbf{95.42}&87.80&\textbf{96.81}&\textbf{70.41}&\textbf{82.73}\\
        \hline
        \end{tabular}
    \caption{Performance comparison on classification datasets given in percentage Top-1 accuracy. \textbf{Bold} values indicate the best performance across all methods for each dataset. Please see Table~\ref{tab:r2_comparison} for the naming convention.} 
    \label{tab:performance_comparison_CLS}
\end{table*}
\begin{table*}[t]
\centering
\small
\begin{tabular}{l | c | c c c | c | c | c c | c c}
\hline
\multirow{2}{*}{\textbf{Method}} 
& \textbf{ADE20K} 
& \multicolumn{3}{c|}{\textbf{COCO}} 
& \textbf{MPII} 
& \textbf{S-COCO $\downarrow$}
& \multicolumn{2}{c|}{\textbf{NYU} $\downarrow$} 
& \multicolumn{2}{c}{\textbf{KITTI} $\downarrow$} \\
\cline{2-11}
& mIoU & AP$^{\mathrm{b}}$ & AP$^{\mathrm{m}}$ & mAP & PCKh & MCE & RMSE & AbsRel & RMSE & AbsRel \\
\hline
iBOT & 52.17 & 0.5158 & 0.4448 & 0.7364 & 0.8697 & 1.76 & 0.3606 & 0.1001 & 2.7469 & 0.0672 \\
+ Ours (all, SE(2)) & 52.38 & 0.5192 & 0.4478 & 0.7371 & \textbf{0.8742} & 1.53 & 0.3514 & 0.0985 & 2.7551 & 0.0671 \\
DINOv2 & 53.49 & 0.5303 & 0.4574 & \textbf{0.7512} & 0.8728 & 1.68 & 0.3454 & 0.0965 & 2.7037 & 0.0664 \\
+ Ours (all, SE(2)) & \textbf{54.12} & \textbf{0.5332} & \textbf{0.4596} & 0.7498 & 0.8736 & \textbf{1.42} & \textbf{0.3413} & \textbf{0.0940} & \textbf{2.6578} & \textbf{0.0640} \\
\hline
\end{tabular}
\caption{Performance comparison on dense prediction datasets. The symbol $\downarrow$ indicates that lower values are better. All the experiments are ran three times and report the mean value.}
\label{tab:performance_comparison_dense}
\end{table*}
\begin{table}[b]
\centering
\small
\begin{tabular}{p{2.2cm} | p{1cm} p{1cm} p{1cm} | p{1cm}}
\hline
\multirow{2}{*}{\textbf{Method}} 
& \multicolumn{3}{c|}{\textbf{DAVIS 2017}} 
& \textbf{VIP} \\
\cline{2-5}
& $\mathcal{J}\&\mathcal{F}_m$ & $\mathcal{J}_m$ & $\mathcal{F}_m$ & mIoU \\
\hline
iBOT & 63.4 & 62.3 & 64.6 & 40.7 \\
+ Ours  & \textbf{65.4} & 63.9 & \textbf{67.0} & \textbf{42.1} \\
DINOv2 & 29.8 & 27.7 & 31.9 & 40.5 \\
+ Ours  & 65.2 & \textbf{64.0} & 66.3 & 41.5 \\
\hline
\end{tabular}
\caption{Performance comparison on the DAVIS and the VIP datasets for video object segmentation. “Ours” corresponds to Ours(all, SE(2)) using all base augmentations and the SE(2) transformation. Results are averaged over three runs.}
\label{tab:davis_performance}
\end{table}

\subsubsection{Linear Probing on Classification Tasks}
We follow the standard SSL evaluation pipeline, where the pretrained network is frozen, and only the linear head is fine-tuned on downstream tasks. The results, as shown in Table \ref{tab:performance_comparison_CLS}, are based on models pretrained on ImageNet1k. We report the performance using Top-1 accuracy, which measures the proportion of test samples for which the model’s most confident prediction matches the ground-truth label.

Surprisingly, we found that our method improved performance compared to the baselines on average and across most datasets and tasks. 
Specifically, Ours(DINOv2) consistently achieved superior performance, with notable improvements in CIFAR10 (98.91\% vs. 98.47\%), CIFAR100 (90.37\% vs. 89.28\%), and Aircraft (71.64\% vs. 70.89\%), surpassing DINOv2 and other baselines. In contrast, only on the Food dataset our method fell slightly behind Ours(iBOT) (88.66\% vs. 87.80\%), which still represented a competitive result. Furthermore, SIE methods, which focus on equivariant features, did not perform well on natural image classification tasks (not shown). As a result, we focused on iBOT and DINOv2-related methods in later experiments.

\subsubsection{Transfer Learning Tasks}
We investigate multiple downstream tasks that leverage equivariant features, including semantic segmentation, object detection, keypoint detection, homography estimation, monocular depth estimation, video object segmentation, semantic part propagation \cite{bhat2023zoedepth,quercia2025enhancing,dam,dam2, davis, vip}. For semantic segmentation, we fine-tune our pretrained model using UPerNet \cite{upernet}. For instance segmentation and object detection, we employ Mask R-CNN \cite{mask-rcnn} with our pretrained model. For homography estimation, we design a 3-layer convolutional head to output the displacement map. For monocular depth estimation we fine-tune our pretrained models using the DepthAnything \cite{dam} pipeline, based on ZoeDepth \cite{bhat2023zoedepth}. Lastly, for video object segmentation and semantic part propagation, we apply our pretrained models using the CropMAE \cite{cropmae} evaluation pipeline.

Table \ref{tab:performance_comparison_dense} shows our method's strong performance across diverse dense prediction tasks on standard benchmarks: semantic segmentation on ADE20k \cite{ZhouCVPR2017}, object detection and instance segmentation on COCO \cite{lin2015microsoftcococommonobjects}, keypoint detection on MPII \cite{AndrilukaCVPR2014}, homography estimation on S-COCO \cite{DeTone2016DeepIH}, monocular depth estimation on NYU \cite{nyu} and KITTI \cite{kitti}, video object segmentation on DAVIS 2017 \cite{davis}, and semantic part propagation on VIP \cite{vip}.

Our DINOv2-based approach consistently improves baseline methods across multiple tasks (but CoCo mAP, where it is on par). For semantic segmentation, we achieve notable improvement on ADE20k (54.12 vs. 53.49 mIoU compared to DINOv2). In homography estimation, our method excels with a Mean Corner Error of 1.42 on S-COCO, substantially better than both iBOT (1.76) and DINOv2 (1.68). For monocular depth estimation, our DINOv2 variant achieves the best RMSE and AbsRel scores on both NYU and KITTI datasets, while our iBOT variant improves on the original iBOT on NYU and matches its performance on KITTI.

For video tasks, our methods show strong improvements. Our iBOT-based approach achieves the best results on both DAVIS 2017 and VIP datasets, while our DINOv2 variant performs comparably to iBOT. Notably, the original DINOv2 performs poorly on DAVIS 2017, but our equivariant approach achieves reasonable performance levels.

These comprehensive results demonstrate that equivariant features effectively enhance performance across computer vision applications, providing consistent improvements over existing state-of-the-art SSL techniques.

\subsection{Comparison with Augmentation-Free Methods}

We compare our approach with reconstruction-based self-supervised learning (SSL) methods that require minimal augmentations, such as MAE \cite{mae}, and those that require no augmentations, such as I-JEPA \cite{i-jepa}. 

\begin{figure}[tb]
    \centering
    \includegraphics[width=0.9\linewidth]{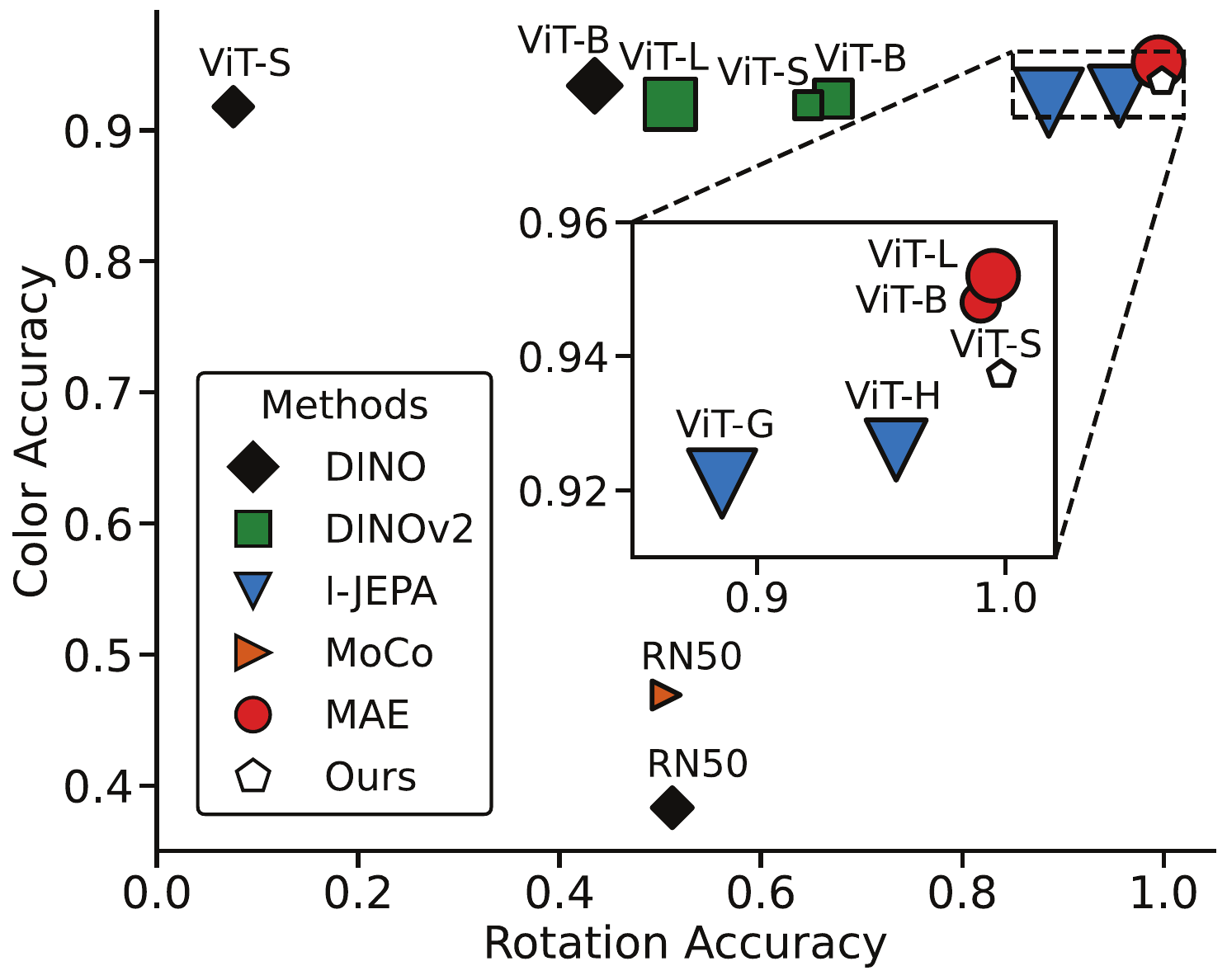}
    \caption{Synthetic tasks results among SSL methods}
    \label{fig:performance_comparison_sota}
\end{figure}

From Figure \ref{fig:performance_comparison_sota}, we observe that all augmentation-based feature-matching methods (DINO \cite{dino}, DINOv2 \cite{Oquab2023DINOv2LR}, MoCo \cite{chen2020mocov2}) perform poorly on rotation prediction tasks, yielding worse results compared to reconstruction-based methods (MAE, I-JEPA). However, our approach enhances the performance of augmentation-based invariance matching methods on both tasks.

We see that the reconstruction-based methods in Figure~\ref{fig:performance_comparison_sota} (MAE, I-JEPA) perform well on transformation prediction tasks. However, they are based on larger models such as ViT-H and ViT-G and/or pretraining on large-scale datasets like ImageNet-22K. In contrast, our approach uses ViT-S and still achieves results comparable to these larger models.

One limitation of reconstruction-based methods is their weaker performance on invariance-related tasks, as shown in Table~\ref{tab:inv_recon_comparison}. 

When evaluated with linear probing, MAE and I-JEPA perform worse than or at best on par with augmentation-based methods iBOT and DINOv2, even though MAE and I-JEPA models are much larger. Our method improves slightly but consistently on the iBOT and DINOv2 baselines with the same model and training data.

We conclude that 
by incorporating transformation reconstruction, our method preserves equivariant representations like other reconstruction approaches, and even slightly outperforms augmentation-based methods on invariant tasks. Thus it combines the best of both worlds.     
\begin{table}[tb]
\centering
\setlength{\tabcolsep}{2pt} 
{\small
\begin{tabular}{l l l c c c}
\hline
\textbf{Method} & \textbf{Pretrain} & \textbf{Arch.} & \textbf{CIFAR100} & \textbf{iNat18} & \textbf{IN} \\
\hline
MAE$\dagger$ & IN1k & H/14 & 77.3 & 32.9 & 77.2 \\
I-JEPA$\dagger$ & IN1k & H/14 & 87.5 & 47.6 & 77.5 \\
I-JEPA$\dagger$ & IN22k & H/14 & 89.5 & 50.5 & 79.3 \\
I-JEPA$\dagger$ & IN22k & G/16 & 89.5 & 55.3 & - \\
iBOT$\ddagger$ & IN1k & L/16 & 87.0 & 57.3 & 81.0 \\
DINOv2$\ddagger$ & IN1k & L/16 & 89.3 & 69.4 & 82.6 \\
DINOv2$\dagger$ & LVD & g/14 & \textbf{94.0} & \textbf{81.6} & \textbf{87.1} \\
\hline
\textbf{Ours (iBOT)} & IN1k & L/16 & \textbf{87.8} & \textbf{58.0} & \textbf{81.6} \\
\textbf{Ours (DINOv2)} & IN1k & L/16 & \textbf{90.4} & \textbf{70.4} & \textbf{82.7} \\
\hline
\end{tabular}
}
\caption{Linear probing performance on invariance tasks compared to models requiring minimal or no data augmentation. $\dagger$ denotes results reported in I-JEPA~\cite{i-jepa}; $\ddagger$ denotes our reproduced pretrained model using the publicly available source code. IN: ImageNet, LVD: LVD-142M, H/14: ViT-H/14, G/16: ViT-G/16, L/16: ViT-L/16, g/14: ViT-g/14}
\label{tab:inv_recon_comparison}
\end{table}

\subsection{Sensitivity Analysis}
\label{subsec:sensitivity-analysis}

Our method involves several hyper-parameters: the split dimension \( d_{\text{equi}} \) of $z^\text{equi}$, i.e. the the portion of the feature vector used to reconstruct the intermediate images $u_k$, the weighting hyper-parameter $\lambda$ in (\ref{eq:total_loss}) for the equivariance-coherent loss $L_\text{recon}$ and the number $K$ of intermediate images for reconstruction. We used iBOT and the small ViT-S/16 for this sensitivity analysis to minimize computational load. Specifically, for the split dimension \( d_{\text{equi}} \) and loss weight \( \lambda \), we performed pretraining for 100 epochs. For transformation analysis, we extended pretraining to 800 epochs. We use SE(2) transformation if not said differently.

\subsubsection{Equivariant Dimension $d_{\text{equi}}$ and Loss Weight $\lambda$} 
We selected all combinations from $\lambda\in\{0.1, 1.0, 5.0\}$ and $d_{\text{equi}}\in\{256, 512, 1024, 2048, 4096\}$ and pretrained on ImageNet-1k as described above. We also tested  \( \lambda > 5.0 \), and observed the training process to become unstable.

\begin{figure}[tb]
    \centering
    \includegraphics[width=0.85\linewidth]{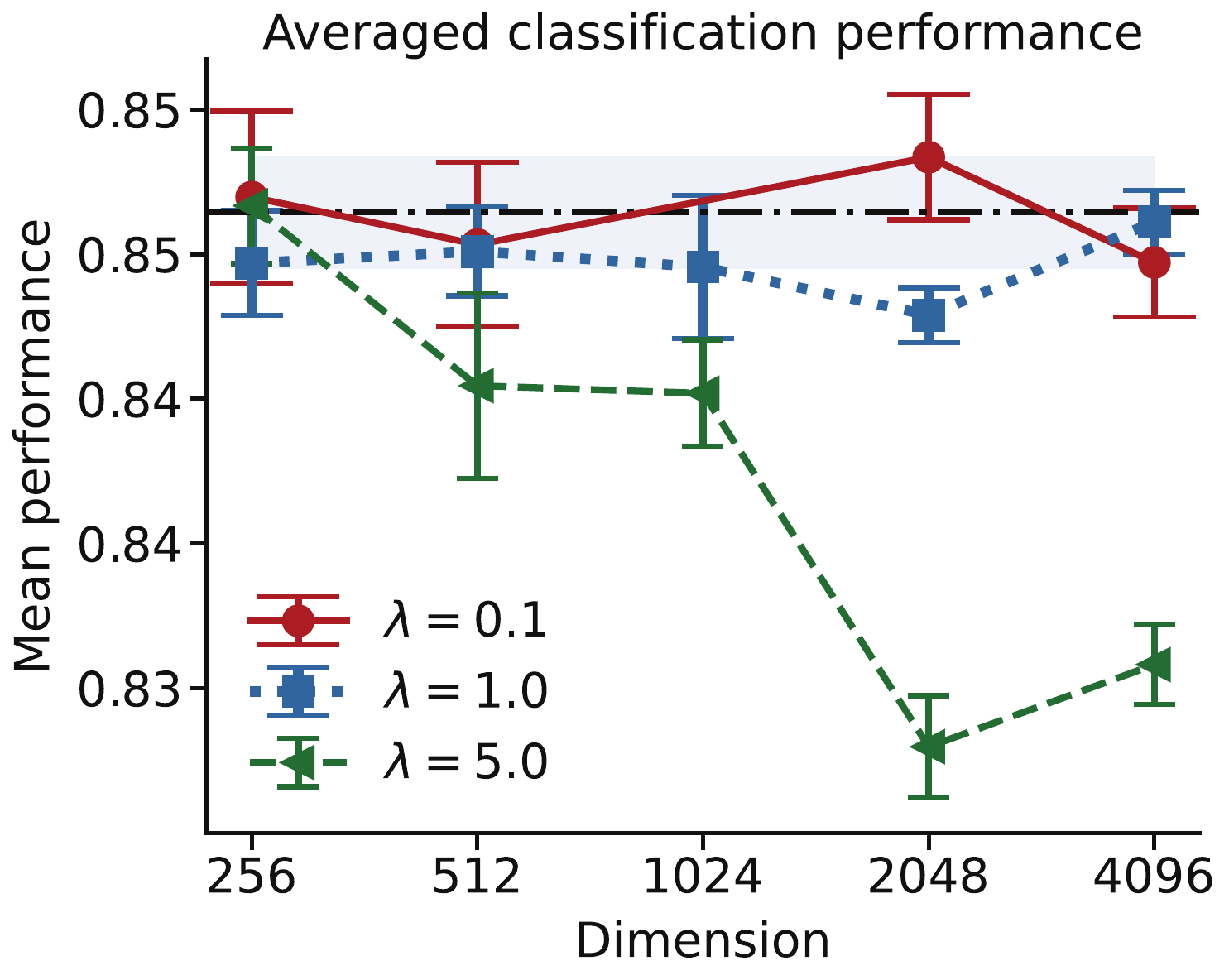}
    \caption{Mean performance across classification tasks.}
    \label{fig:inv_task_mean}
\end{figure}

\begin{figure}[tb]
    \centering
    \includegraphics[width=0.85\linewidth]{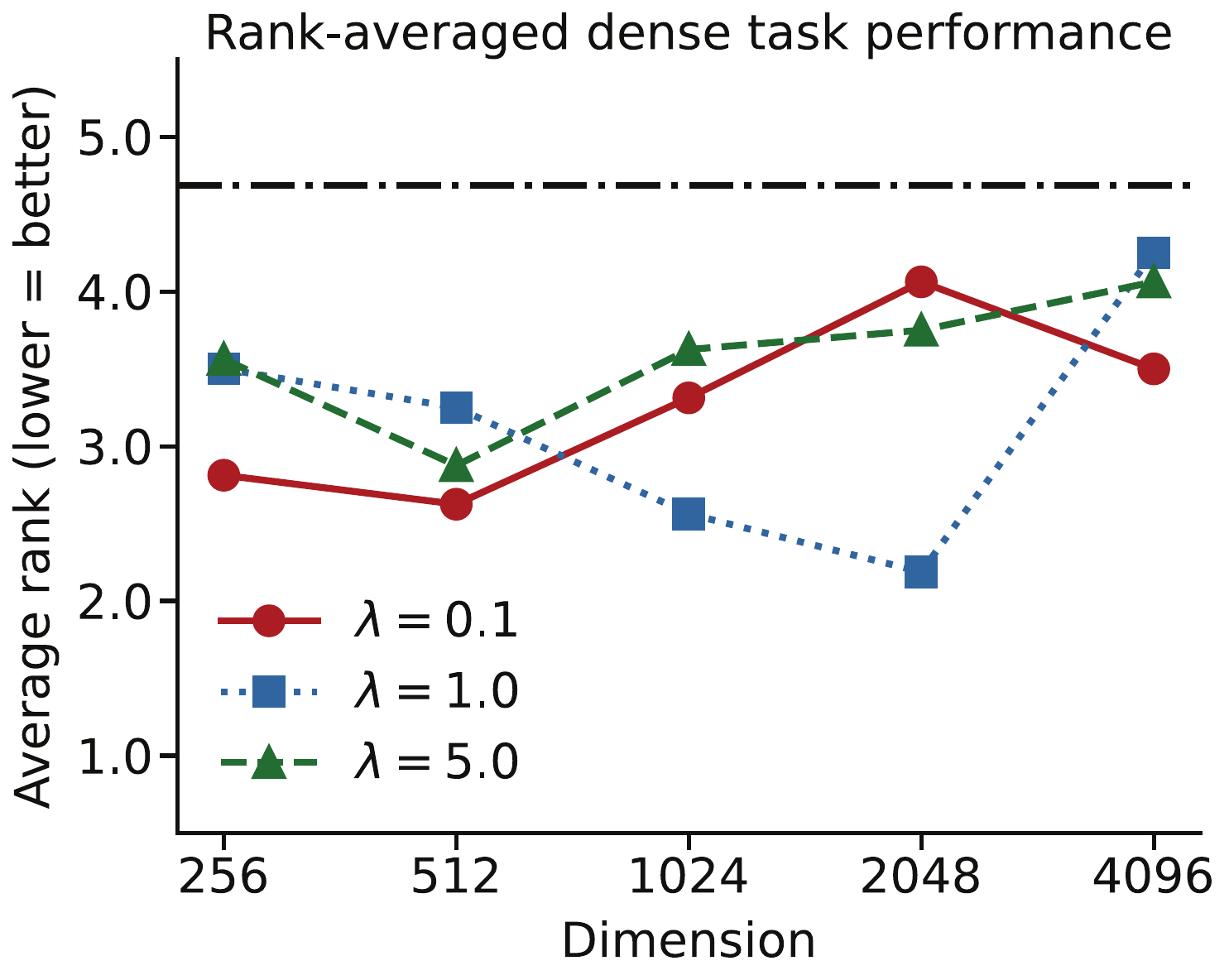}
    \caption{Mean performance across dense prediction tasks.}
    \label{fig:equiv_task_mean}
\end{figure}

For the classification tasks, we provide in  Figure~\ref{fig:inv_task_mean}  the mean and standard error of the accuracy for the hyper-parameter combination. For dense prediction tasks with different performance measures, we provide in  Figure~\ref{fig:equiv_task_mean}   the average rank for each hyper-parameter combination. For the classification tasks we observe for small $\lambda=0.1$ and medium $\lambda=1$ weighting parameter a rather stable behaviour of the mean accuracy around the baseline (dashed line) performance. For larger weighting parameter $\lambda=1$ the performance decreases with larger portion of the feature vector used for the intermediate reconstruction task. The overall observation is reasonable, as classification tasks do not benefit from equivariance as dense prediction tasks do. Thus, increasing the weight parameter, i.e. focusing on the equivariant reconstruction task while giving less space portion of the feature vector for invariance, i.e. increasing $d_{\text{equi}}$, leads to a significant performance decrease. The qualitative behavior of our approach for the dense prediction tasks is different as, on one hand,  our method outperforms the baseline for all hyper-parameters with respect to the average rank and on the other hand we observe an optimal parameter combination at \( \lambda = 1 \) and \( d_{\text{equi}} = 2048 \). 

\subsubsection{Number of Inbetween Images}
In Table \ref{tab:num_images}, we investigate the effects of the number of in-between images. We observe that with only one in-between image, the performance is sub-optimal. Increasing the number of intermediate images to two significantly improves performance on synthetic tasks. Notably, adding more than two in-between images or incorporating the augmented view $v_2$ for reconstruction does
not lead to further improvements. Considering both performance gains and GPU memory constraints, we select two in-between images for our experiments.
\begin{table}[tb]
\centering
\small
\begin{tabular}{lcccc}
\hline
\# of Images & $R^2(\text{rot})$ & $R^2(\text{color})$ & $R^2(\text{blur})$ & $R^2(\text{trans})$ \\
\hline
1 & 0.2915 & 0.4708 & 0.8725 & 0.4230 \\
2 & \textbf{0.9983} & \textbf{0.9373} & \textbf{0.9689} & \textbf{0.9801} \\
3 & 0.9981 & 0.9215 & 0.9506 & 0.9795 \\
2 + final & 0.9981 & 0.9311 & 0.9562 & 0.9771 \\
\hline
\end{tabular}
\caption{Number of Inbetween images investigation. The model is Ours(VICReg, all, rot), cmp. Table~\ref{tab:r2_comparison}.}
\label{tab:num_images}
\end{table}

\section{Summary and Conclusions}

We propose a novel approach for augmentation-based SSL methods that enhances the learning of equivariant-coherent features by reconstructing intermediate representations of transformed images. Our method significantly boosts performance on equivariance-focused synthetic tasks and surpasses competitors like SIE. Moreover, we achieve comparable or superior results on real-world imaging tasks using iBOT and DINOv2 as base methods. This approach provides a promising direction for improving the generalization of SSL methods and can be easily adapted to other SSL frameworks.

\section*{Acknowledgments}
The authors gratefully acknowledge the Gauss Centre for Supercomputing e.V. (www.gauss-centre.eu) for funding this project by providing computing time on the GCS Supercomputer JUWELS \cite{JUWELS} at Jülich Supercomputing Centre (JSC).
\bibliography{aaai2026}

\end{document}